\documentclass[letterpaper,10pt]{IEEEtran}

\usepackage{multirow,amssymb,amsmath}
\usepackage{graphicx}
\usepackage{wasysym}
\begin{document}
\title{A Novel Windowing Technique for Efficient Computation of MFCC for Speaker Recognition}

\author{Md~Sahidullah,~\IEEEmembership{Student Member,~IEEE,}
        Goutam~Saha,~\IEEEmembership{Member,~IEEE,}
\thanks{The authors are with the Department
of Electronics and Electrical Communication Engineering, Indian Institute of Technology Kharagpur, Kharagpur, West Bengal,
721302, India. e-mail: sahidullahmd@gmail.com; gsaha@ece.iitkgp.ernet.in}}

\maketitle

\begin{abstract}
In this paper, we propose a novel family of windowing technique to compute Mel Frequency Cepstral Coefficient (MFCC) for automatic speaker recognition from speech. The proposed method is based on fundamental property of discrete time Fourier transform (DTFT) related to differentiation in frequency domain. Classical windowing scheme such as Hamming window is modified to obtain derivatives of discrete time Fourier transform coefficients. It has been mathematically shown that the slope and phase of power spectrum are inherently incorporated in newly computed cepstrum. Speaker recognition systems based on our proposed family of window functions are shown to attain substantial and consistent performance improvement over baseline single tapered Hamming window as well as recently proposed multitaper windowing technique.
\end{abstract}


\section{Introduction}
\PARstart{M}el frequency cepstum coefficient (MFCC) extraction schemes use discrete Fourier transform (DFT) for calculating short-term power spectrum of speech signal. During this process, Hamming or Hanning window is applied to raw speech frames in order to reduce spectral leakage effect. These windows have reasonable sidelobe and mainlobe characteristics which are required for DFT computation. However, there exists various other window functions which also have good behavior in terms of certain parameters of their frequency responses~\cite{HarrisWindow}. In practice, selecting the optimal window function for speech processing application is still an open challenge~\cite{SmithAudio}. Recently, alternatives of Hamming window have drawn attention of the researchers~\cite{Mottaghi, WangY}. For example, performance of speaker recognition systems based on MFCC, extracted using multitaper window function, are shown comparatively robust than existing single tapered Hamming window based approach~\cite{SandbergKinnunen}.
\par
In this work, we propose a simple time domain processing of speech after it is multiplied with a standard window. The processing is based on well-known \emph{difference in frequency} property of discrete time Fourier transform~\cite{OppenSignals}, and it can be easily integrated with standard window during DFT computation. Due to the proposed modification, we inherently compute derivative of Fourier transform. Power spectrum is computed from those differentiated Fourier coefficients. There are evidences that speaker discriminating attribute is present in slope of power spectrum~\cite{SahidullahBlockTransform} as well as in phase information~\cite{NakagawaPhase}. In this paper, we have mathematically shown that our proposed technique integrates both slope and phase information with magnitude spectrum. Therefore, it can be hypothesized that the speech feature extraction from these modified Fourier coefficients will give better recognition performance. We have evaluated the performance in multiple databases for speaker verification (SV) task, and consistent performance improvement is achieved over Hamming window based baseline system.
\par
The rest of the paper is organized as follows. In Section~\ref{Section:Proposed}, we describe the proposed windowing scheme and its features. In addition to that, the effect of newly introduced window in power spectrum computation is mathematically analyzed. Experimental results are shown in Section~\ref{Section:Experiment}. Finally, the paper is concluded in Section~\ref{Section:Conclusion}.

\section{Proposed Windowing Method}\label{Section:Proposed}

\subsection{Design of proposed window function}

Let $x(n)$ be a windowed speech frame of length $\mathit{N}$  and its DTFT is given by, $X(e^{j\omega})$. We know from differentiation in frequency property~\cite{OppenSignals} that DTFT of $nx(n)$ can be written as,

\begin{equation}
\hat{X}(e^{j\omega})=j\frac{{dX(e^{j\omega } )}}{{d\omega }}.
\label{Eq1}
\end{equation}

As DFT coefficients $X(k)$ are samples of DTFT at $\omega=\frac{2\pi k}{N}$, DFT of $nx(n)$ are discrete samples of $\hat{X}(e^{j\omega})$ at $\omega=\frac{2\pi k}{N}$. Therefore, $\hat{X}(k)=\hat{X}(e^{j\omega})|_{\omega=\frac{2\pi k}{N}}$ are the DFT coefficients of $nx(n)$.

Since $x(n)$ is a windowed speech frame, it can be represented as $w(n)s(n)$, where $s(n)$ is raw speech frame and $w(n)$ is window function. We propose new window function as $\hat{w}(n)=nw(n)$. The windowed speech frame is then represented as $\hat{x}(n)=\hat{w}(n)s(n)$.

\begin{figure}[t]
  \centerline{\includegraphics[width=9cm,height=6.5cm]{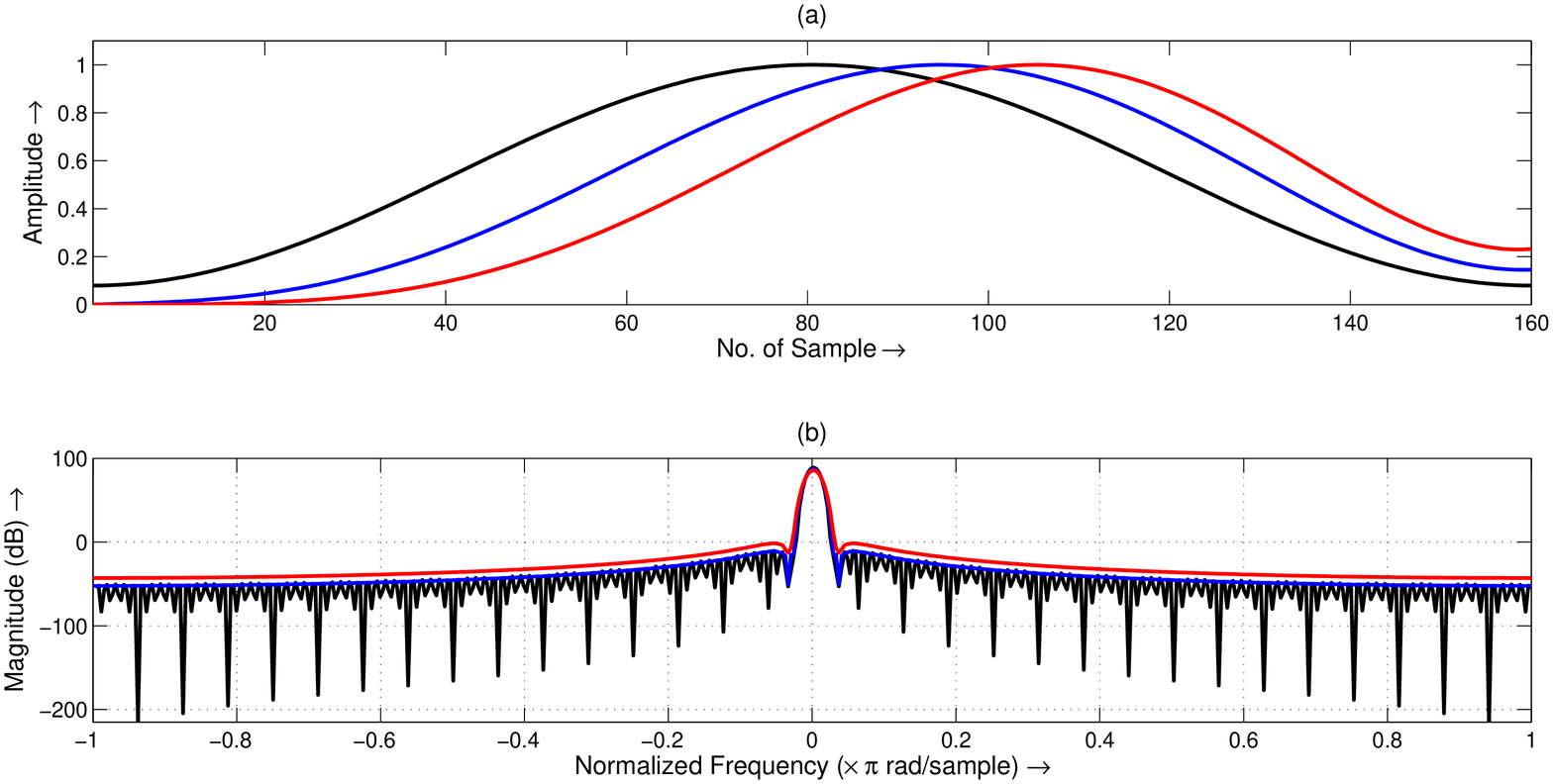}}
  \caption{{\it Comparison of Hamming window (black) with first (blue) and second (red) order differentiation based window in (a)time domain and (b)frequency domain for a window of size $160$ samples. Amplitude of all the window functions are normalized to one for visual clarity.}}
  \label{Fig:Window}
\end{figure}

From generalization of differentiation in frequency property, we can write that, for an integer $\tau$, DTFT of $n^{\tau}x(n)$ is $j^{\tau}\frac{{d^{\tau}X(e^{j\omega } )}}{{d^{\tau}\omega }}$. Therefore, the proposed window function of $\tau$-th order window can be written as $n^{\tau}w(n)$. Standard Hamming window can be viewed as \emph{zero order window} of proposed family. The window functions are shown in Fig.~\ref{Fig:Window} for first and second order along with Hamming window. Note that in contrast to frequently used window functions, the newly introduced family of window functions is asymmetric and non-tapered.

\subsection{Characteristics of the proposed window function}
Commonly, the effectiveness of an window function is judged by different performance metrics~\cite{HarrisWindow}. In order to evaluate the performance of the window in DFT computation, various performance metrics are computed prior to the application of this window function in speech feature extraction. We have calculated three widely used performance evaluation metrics: spectral leakage factor, relative side lobe attenuation, and mainlobe width ($-3$dB) of the Hamming and proposed windows of different orders. The results are shown in Table~\ref{tablewindow} for window size of $160$ samples. It can be observed that with the increase of order, the spectral leakage increases and sidelobe attenuation decreases to some extent which have minor effect in recognition performance. However, considerable increase in mainlobe width will help to estimate smooth power spectrum, and that is expected to improve recognition performance~\cite{DSPProakisManolakis}.

\begin{table} [t,h]
\caption{\label{tablewindow} {\it Performance metrics of various window functions. Sequence length is $160$ samples i.e. $20$ms for sample rate $8$kHz.}}
\vspace{2mm}
\centerline{
\begin{tabular}{|c|c|c|c|}
\hline
\multirow{2}{*}{Window} & Leakage  & Relative              &     Mainlobe  \\
                        & Factor   & Sidelobe Attenuation  &      Width \\
\hline     \hline
Hamming    &  0.04\%   &  -42.6 dB &  0.015625\\
\hline
$\tau=1$   &  0.06\%   &  -42.6 dB &  0.017578\\
\hline
$\tau=2$   &  0.17\%   &  -37.9 dB &  0.018555\\
\hline
\end{tabular}}
\end{table}

\subsection{Effect of the proposed window in power spectrum computation}
In this subsection, we find out a mathematical connection between power spectrum of proposed windowed speech frame and power spectrum of original Hamming windowed speech frame.
\par
Let us assume that power spectrum of  Hamming windowed signal is given by $P(\omega)$, and power spectrum of the proposed window is $\hat{P}(\omega)$. Therefore, $P(\omega)=H^{2}(\omega)=\left| {X(e^{j\omega } )} \right|^{2}$ and $\hat{P}(\omega)=\hat{H}^{2}(\omega)=\left| {\frac{{dX(e^{j\omega } )}}{{d\omega }}} \right|^2$, where $H(\omega)$ and $\hat{H}(\omega)$ are magnitude spectrum of two signals respectively. Now, since ${X(e^{j\omega})}$ can be decomposed into a real, $X_{R}(\omega)$ and imaginary, $X_{I}(\omega)$ part, the slope of magnitude spectrum of Hamming windowed speech signal can be written as,

\begin{eqnarray}
\frac{{dH(\omega)}}{{d\omega }} &=& \frac{{d\left| {X(e^{j\omega } )} \right|}}{{d\omega }} \nonumber \\
&=& \frac{{X_R(\omega) }}{{\sqrt {X_R^2(\omega)  + X_I^2(\omega) } }}\frac{{dX_R(\omega) }}{{d\omega }} \nonumber\\
&&+ \frac{{X_I(\omega) }}{{\sqrt {X_R^2(\omega)  + X_I^2(\omega) } }}\frac{{dX_I(\omega) }}{{d\omega }}
\label{Eq3}
\end{eqnarray}


On the other hand, magnitude spectrum of the modified signal can be written as,

\begin{eqnarray}
\hat{H}(\omega ) = \left| {\frac{{dX(e^{j\omega } )}}{{d\omega }}} \right| = \sqrt {\left( {\frac{{dX_R }}{{d\omega }}} \right)^2  + \left( {\frac{{dX_I }}{{d\omega }}} \right)^2 }
\label{Eq4}
\end{eqnarray}

Now, if we consider that $ {\frac{{dX_R(\omega) }}{{d\omega }}}  = a(\omega)\cos \varphi(\omega)$ and $ {\frac{{dX_I(\omega) }}{{d\omega }}}  = a(\omega)\sin \varphi(\omega)$, then $a(\omega)=\sqrt {\left( {\frac{{dX_R(\omega) }}{{d\omega }}} \right)^2  + \left( {\frac{{dX_I(\omega) }}{{d\omega }}} \right)^2 }$ and $\varphi(\omega)  = \tan ^{ - 1} \frac{{\frac{{dX_I(\omega) }}{{d\omega }}}}{{\frac{{dX_R(\omega) }}{{d\omega }}}} = \tan ^{ - 1} \frac{dX_I(\omega)}{dX_R(\omega)}$.

Therefore, from Equation (~\ref{Eq4}),

\begin{equation}
a(\omega)=\hat{H}(\omega)
\label{Eq5}
\end{equation}

On the other hand, if we put $\frac{{X_R(\omega) }}{{\sqrt {X_R^2(\omega)  + X_I^2(\omega) } }} = \cos \phi (\omega)$ and $\frac{{X_I(\omega) }}{{\sqrt {X_R^2(\omega)  + X_I^2(\omega) } }} = \sin \phi (\omega)$ in Equation (~\ref{Eq3}) we get,

\begin{eqnarray*}
\frac{{dH(\omega )}}{{d\omega }} &=& \frac{{X_R(\omega) }}{{\sqrt {X_R^2(\omega)  + X_I^2(\omega) } }} \times a(\omega)\cos \varphi(\omega) \nonumber \\
&& \quad\frac{{X_I(\omega) }}{{\sqrt {X_R^2(\omega)  + X_I^2(\omega) } }} \times a(\omega)\sin \varphi(\omega) \nonumber \\
&=& a(\omega) \cos \phi (\omega) \cos \varphi(\omega) \nonumber \\
&& \quad +a(\omega) \sin \phi (\omega) \sin \varphi(\omega) \nonumber \\
&=& a(\omega) \cos [\phi (\omega)-\varphi(\omega)]
\label{Eq61}
\end{eqnarray*}

\begin{equation}
\therefore \frac{{dH(\omega )}}{{d\omega }} = a(\omega) \cos [\phi (\omega)-\varphi(\omega)]
\label{Eq62}
\end{equation}

where $\phi (\omega)  = \tan ^{-1} \frac{X_I(\omega)}{X_R(\omega)}$.

Therefore, from Equation (~\ref{Eq5}) and Equation (~\ref{Eq62}), we get,

\begin{equation}
\hat H(w) = \frac{{dH(\omega )}}{{d\omega }} \times \sec \left[\phi(\omega)-\varphi(\omega)\right]
\label{Eq7}
\end{equation}

Finally, we can write the final expression of the output power spectrum $\hat{P}(\omega)$ as,

\begin{equation}
\hat{H}^{2}(\omega)=\frac{1}{4P(\omega)}\left[{\frac{{dP(\omega )}}{{d\omega }}}\right]^{2} \times \sec^{2} \left[\phi(\omega)-\varphi(\omega)\right].
\label{Eq8}
\end{equation}

\[
 {}
\]

The term $\frac{{dP(\omega )}}{{d\omega }}$  in Equation (~\ref{Eq8}) corresponds to the slope of the power spectrum of the Hamming windowed speech at frequency $\omega$. Hence, as a consequence of power spectrum computation from derivative of fourier transform, we obtain a modified power spectrum which is related to the slope of original power spectrum. Apart from it, the newly formulated power spectrum is also related to phase spectrum of the signal $\phi(\omega)$. Using a more complicated computation, it can also be shown that the higher order version of proposed differentiation window (e.g. for $\tau>1$) will compute power spectrum with higher order derivative of $P(\omega)$.

The modified DFT magnitude coefficients are nothing but the samples of $\hat{H}(\omega)$  at $\omega=\frac{2 \pi k}{N}$. Therefore, mel cepstrum computation using proposed window integrates the slope of power spectrum, phase, and of course, power spectrum of the signal. It is expected that the speech feature will be more efficient compared to the standard cepstrum which is solely based on power spectrum.

\section{Experimental Setup and Results}\label{Section:Experiment}

\subsection{Speaker Recognition Setup}
\subsubsection{Database}
SV experiments are conducted on multiple large population NIST corpora for obtaining statistically significant results. We have chosen SRE $2001$, SRE $2004$, and SRE $2006$. The database descriptions for current experiments are briefly shown in Table~\ref{tabledatabase}.

\begin{table} [t,h]
\caption{\label{tabledatabase} {\it Database description (coretest section) for the performance evaluation of various window functions.}}
\vspace{2mm}
\centerline{
\begin{tabular}{|c|c|c|c|}
\hline
 & SRE 2001  & SRE 2004              &     SRE 2006  \\
\hline     \hline
Target Models      &  74\male, 100\female &  246\male, 370\female  &  354\male, 462\female \\
\hline
Test Segments                  &  2038       &  1174        &  3735 \\
\hline
Total Trial                    &  22418      &  26224       &  51068 \\
\hline
True Trial                     &  2038       &  2386        &  3616\\
\hline
Impostor Trial                 &  20380      &  23838       &  47452\\
\hline
\end{tabular}}
\end{table}

\subsubsection{Feature Extraction}
MFCC features have been extracted for different types of window functions. $38$ dimensional feature vectors are computed using $20$ filters linearly spaced in Mel scale from speech frames of size $20$ms (with $50\%$ overlap). Detailed explanation of used MFCC computation technique is available in~\cite{SahidullahBlockTransform}.

\subsubsection{Classifier Description}
State-of-the art speaker recognition system uses Gaussian mixture model-universal background model (GMM-UBM) based classifier~\cite{Reynoldsadapted}. The speech data for UBM training are taken from development data of SRE $2001$ and training section of SRE $2003$ for the evaluation of SRE $2001$ and SRE $2004$ respectively. Number of mixtures are set at $256$ for these experiments. Here, gender dependent GMM clusters are initialized using binary split based vector quantization. The final UBM parameters are estimated using EM algorithm. Target models are created by adapting only the means of the UBM with relevance factor $14$. During the score computation, top-$5$ Gaussians of corresponding background model per each frame are considered.
\par
For the evaluation of SRE $2006$, the GMM-UBM system is trained with $512$ mixtures of gender dependent UBM with complete one side training data of SRE $2004$ (i.e. $246$ male and $370$ female utterances). $zt$-score normalization is performed on raw score of GMM-UBM system. Normalization data is obtained from one side section of SRE $2004$. Experiments are also conducted using classifiers based on GMM supervector and support vector machine (GSV-SVM)~\cite{CampbellGLDS}. This is based on the same UBM of GMM-UBM system. The negative examples of SVM are obtained from the same data used for UBM preparation. Experiments are also carried out with nuisance attribute projection (NAP) based channel compensation technique~\cite{CampbellGLDSNAP}. Channel factors are obtained using the speech signals of SRE $2004$. All together, $699$ utterances of $101$ male and $905$ utterances of $142$ female are utilized to train the NAP projection matrix of co-rank $64$.

\begin{table*} [t,h]
\caption{\label{tablesre01sre04} {\it SV performance on NIST SRE 2001 and NIST SRE 2004 using various window functions for GMM-UBM based system.}}
\vspace{2mm}
\centerline{
\begin{tabular}{|c|c|c|c|c|}
\hline
Window & \multicolumn{2}{|c|}{NIST SRE 2001} & \multicolumn{2}{|c|}{NIST SRE 2004}\\
\cline{2-5}
Type                 & EER (in \%) & minDCF $\times$ 100  & EER (in \%) & minDCF $\times$ 100  \\
\hline               \hline
Hamming              & 8.2434   & 3.5763  & 14.9629  &  6.3231  \\
\hline
Multitaper ($k=6$)   & 8.0471   & 3.5778  & 15.2501  &  6.4363  \\
\hline
Multitaper ($k=12$)  & 10.9372  & 4.6606  & 18.0196  &  7.2271  \\
\hline
First Order          & 8.1943   & 3.5672  & 14.6694  &  6.2501  \\
\hline
Second Order         & 7.6055   & 3.3763  & 14.3255  &  6.1050  \\
\hline
\end{tabular}}
\end{table*}

\subsection{Results}
\begin{figure*}[!htp]
\centerline{\includegraphics[width=18cm]{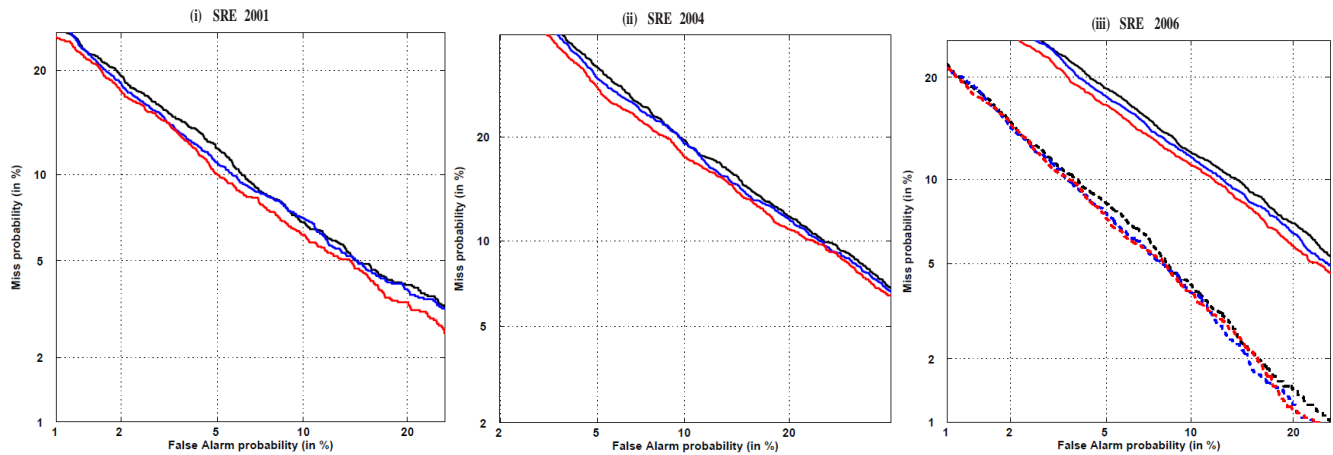}}
\caption{{\it DET plots of different window based systems (Black: Hamming, Blue: first order, Red: second order) are shown for (i)SRE 2001, (ii)SRE 2004, (iii)SRE 2006. In subfigure (iii), the dotted lines show results for GSV-SVM system with NAP.}}
\label{DETPLOT}
\end{figure*}

Speaker recognition experiments are carried out with different window function keeping other blocks identical i.e. pre-processing, feature extraction and classification are precisely same for all various window based systems. We first evaluate the performance on SRE $2001$ and SRE $2004$ with classical GMM-UBM system. The performance of proposed windows (first and second order) are compared with single tapered Hamming window as well as recently proposed multitaper window. The performance has been evaluated with multipeak taper of size (denoted by $k$ in Table~\ref{tablesre01sre04}) $6$ and $12$ as mentioned in~\cite{Kinnulow, SandbergKinnunen}. The results are shown in Table~\ref {tablesre01sre04} and corresponding detection error trade-off (DET) plots are shown in Fig.~\ref{DETPLOT}(i) and Fig.~\ref{DETPLOT}(ii). Equal error rate (EER) and minimum detection cost function
(minDCF) of SV systems based on newly proposed window functions are consistently better for both the databases. In comparison with baseline Hamming window based system, we have obtained $0.6\%$ and $7.74\%$ relative improvement in EER, and $0.26\%$ and $5.59\%$ relative improvement in minDCF for SRE $2001$. In contrast, for SRE $2004$, the relative improvements in EER are $1.96\%$ and $4.26\%$, and for minDCF these are $1.15\%$ and $3.45\%$. Interestingly, we have observed that multitaper windowing techniques do not give better performance as compared to proposed method.

In Table~\ref{tablesre06}, the performance is shown for different classifiers on SRE $2006$. Also, in this case, we have achieved consistent and reasonable performance improvement for proposed window based SV system. The DET plot is shown in Fig.~\ref{DETPLOT}(iii) for both GMM-UBM and GSV-SVM (with NAP) system. We can easily interpret from the curves that SV system based on the proposed window functions are consistently better than Hamming window based baseline system. It is also observed that performances of second order window based systems are better than first order window based system.

\begin{table*} [t,h]
\caption{\label{tablesre06} {\it SV performance using various systems for different window function.}}
\vspace{2mm}
\centerline{
\begin{tabular}{|c|c|c|c|c|c|c|}
\hline
Window & \multicolumn{2}{|c|}{GMM-UBM} & \multicolumn{2}{|c|}{GSV-SVM}  & \multicolumn{2}{|c|}{GSV-SVM (with NAP)}\\
\cline{2-7}
Type & EER (in \%) & minDCF $\times$ 100 & EER (in \%) & minDCF $\times$ 100 & EER (in \%) & minDCF $\times$ 100  \\
\hline     \hline
Hamming             &  11.4493  & 4.4702   & 8.8471  & 4.0330 & 6.6419   & 3.1161  \\
\hline
Multitaper ($k=6$)  &  11.6981  & 4.5493   & 9.0705  & 4.2211 & 6.8886   & 3.2725  \\
\hline
Multitaper ($k=12$) &  14.2971  & 5.2299   & 11.430  & 5.0286 & 8.2416   & 3.9699  \\
\hline
Proposed  First Order         & 10.9856   & 4.3521   & 8.3792  &  4.0233 & 6.2503 & 3.0961  \\
\hline
Proposed  Second Order        & 10.7559   & 4.2627 & 8.3242  & 3.9555 &  6.1359      &  3.0646 \\
\hline
\end{tabular}}
\end{table*}

\section{Conclusion}\label{Section:Conclusion}
In this paper, we have focused on the usage of a class of window functions by which more effective speech feature can be computed. The newly formulated feature represents the power spectrum of the original spectrum as well as its derivative. In addition to that, it also integrates phase information which is also relevant for speaker recognition. Speaker recognition system based on proposed windowing schemes are evaluated on different NIST databases. We have achieved consistent performance improvement over baseline Hamming window based technique on various combinations of classifiers and databases.

\bibliographystyle{IEEEtran}
\bibliography{latexbib}

\end{document}